# Nonparametric Divergence Estimation with Applications to Machine Learning on Distributions

**Barnabás Póczos**
School of Computer Science
Carnegie Mellon University
Pittsburgh, PA
USA, 15213

**Liang Xiong**
School of Computer Science
Carnegie Mellon University
Pittsburgh, PA
USA, 15213

**Jeff Schneider**
School of Computer Science
Carnegie Mellon University
Pittsburgh, PA
USA, 15213

## Abstract

Low-dimensional embedding, manifold learning, clustering, classification, and anomaly detection are among the most important problems in machine learning. The existing methods usually consider the case when each instance has a fixed, finite-dimensional feature representation. Here we consider a different setting. We assume that each instance corresponds to a continuous probability distribution. These distributions are unknown, but we are given some i.i.d. samples from each distribution. Our goal is to estimate the distances between these distributions and use these distances to perform low-dimensional embedding, clustering/classification, or anomaly detection for the distributions. We present estimation algorithms, describe how to apply them for machine learning tasks on distributions, and show empirical results on synthetic data, real word images, and astronomical data sets.

## 1 Introduction

Consider the following problem where we have several independent groups of people, and the groups might have different size. In each group we make some measurements, for example we measure the blood pressures of a few people. Suppose that in each group there is a well-defined distribution of blood pressure, and each measurement is an i.i.d. sample from this distribution. The question we want to study is how different these groups are from each other. In particular, is it possible to arrange the groups into some natural clusters using the measurements? Can we embed the distributions (i.e. the groups) into a small dimensional space preserving proximity where they would reveal some structure? Can we detect interesting, unusual groups? It can happen that each measurement in a group looks normal, that is the blood pressure values are in the same normal range, but the distribution might be different from the distributions in the other groups. Can we detect these anomalous groups? The standard anomaly/novelty detection methods only focus on finding individual points (Chandola et al., 2009). Our group anomaly detection task, however, is different; we want to find anomalous groups of points in which each individual point can be normal.

Similar questions arise in many other scientific research areas. Contemporary observatories, such as the Sloan Digital Sky Survey, produce a vast amount of data about galaxies and other celestial objects. It is an important question how to find anomalous clusters of galaxies, where each galaxy in the cluster is normal, but the cluster members together exhibit unusual behavior, i.e., the distribution of the feature vectors in the cluster is different from the distributions of the feature vectors in other clusters, although each feature vector is normal.

Low-dimensional embedding and manifold learning (Roweis and Saul, 2000) are well studied problems; several different algorithms have been proposed for this problem (Borg and Groenen, 2005, Tenenbaum et al., 2000, Sun et al., 2010, Zhang and Zha, 2004, Belkin and Niyogi, 2003, Donoho and Grimes, 2003). These methods usually consider a fixed dimensional feature representation and try to embed these feature vectors into a lower dimensional space. In this paper we generalize this problem and propose a method that is able to embed *distributions* into a lower dimensional space. In this case the original large dimensional space is the space of distributions. In contrast to standard manifold learning problems, here the original large dimensional instances (i.e. distributions) are not known either, only a few i.i.d. samples are given from them. Our goal is to embed them into a lower dimensional space without estimating their densities.

Clustering and classification are also among the most frequent machine learning problems. The most well-known algorithms can only deal with fixed, finite-dimensional representations, and they are not developed to work on distributions. We will show how these problems can be solved using our methods.

To study these kind of questions we need to measure the distance between distributions. We will use the $L_2$ distance and the Rényi divergence for this purpose. While the question of how far distributions are from each other is an important and very basic statistical problem, interestingly, we know very little about how to estimate it efficiently. If the distributions are Gaussian mixtures, then there is a closed form expression for the $L_2$ divergence between them. Nonetheless, we do not have closed form expression for Rényi, Kullback-Leibler, or many other divergences.

Several different probability divergences have been defined in the literature, but only a few papers look for efficient estimation methods for them. Wang et al. (2009b)[1] and Pérez-Cruz (2008)[2] provided an estimator for KL-divergence, and Póczos and Schneider (2011) developed nonparametric methods for the Rényi and Tsallis divergence estimation. Hero et al. (2002a,b) also investigated the Rényi divergence estimation problem but assumed that one of the two density functions is known. Gupta and Srivastava (2010) developed algorithms for estimating the Shannon entropy and the KL divergence for certain parametric families. Recently, Nguyen et al. (2009, 2010) developed methods for estimating $f$-divergences using their variational characterization properties. They estimate the likelihood ratio of the two underlying densities and plug that into the divergence formulas. This approach involves solving a convex minimization problem over an infinite-dimensional function space. For certain function classes defined by reproducing kernel Hilbert spaces (RKHS), however, they were able to reduce the computational load from solving infinite-dimensional problems to solving $n$-dimensional problems, where $n$ denotes the sample size. When $n$ is large, solving these convex problems can still be very demanding. Furthermore, choosing an appropriate RKHS also introduces questions regarding model selection.

In this paper we will use Rényi and $L_2$ divergence estimators. They do not need to solve minimization problems over function classes; we only need to calculate certain $k$-nearest-neighbor ($k$-NN) based statistics. The estimators are consistent, but do not require estimating the densities. Recently, Sricharan et al. (2010) proposed $k$-nearest-neighbor based methods for estimating non-linear functionals of density, but in contrast to our approach, they were interested in the case where $k$ increases with the sample size.

For the estimation of Rényi divergence, we will use the estimator developed in Póczos and Schneider (2011). To estimate the $L_2$ divergence, we will borrow some ideas from Leonenko et al. (2008) and Goria et al. (2005), who considered Shannon and Rényi-$\alpha$ entropy estimation from a single sample.[3] In contrast, we propose divergence estimators using two independent samples. Recently, Póczos et al. (2010), Pál et al. (2010) proposed a method for consistent Rényi information estimation, but this estimator also uses one sample only and cannot be used for estimating divergences. Further information and useful reviews of several different divergences can be found, e.g., in Villmann and Haase (2010), Cichocki et al. (2009), and Wang et al. (2009a).

The main contribution of our work is to propose new algorithms for (i) the low dimensional embedding of distributions, (ii) for the group anomaly detection, and (iii) for the clustering/classification of distributions problems.

The paper is organized as follows. In the next section we review the Rényi, and $L_2$ divergences. In section 2 we formally define our problem. Section 3 introduces our proposed divergence estimators, and here we also summarize their theoretical results about asymptotic unbiasedness and consistency. Due to the lack of space we can present only sketches of the consistency proofs (Section 4); the details will be published elsewhere. Section 5 contains the results of our numerical experiments that demonstrate the applicability of the estimators for group anomaly detection, low-dimensional embedding, clustering and classification. Finally, we conclude with a discussion of our work.

## 2 Formal Problem Definition

In this section we briefly review the Rényi-$\alpha$ (Rényi, 1961) and $L_2$ divergences, and then formally define the goal of our paper.

**Definition 1.** *Let $p$ and $q$ be densities over $\mathbb{R}^d$, and $\alpha \in \mathbb{R} \setminus \{1\}$. The* Rényi-$\alpha$ divergence *is defined as*

$$R_\alpha(p\|q) \doteq \frac{1}{\alpha - 1} \log \int p^\alpha(x) q^{1-\alpha}(x)\, \mathrm{d}x. \quad (1)$$

---

[1] The proposed estimator is consistent, but there is an apparent error in their proofs; they applied the reverse Fatou lemma under conditions when it does not hold. It is not obvious how this portion of the proof can be remedied.

[2] The consistency proof of this paper also has some errors: the author applies the strong law of large numbers under conditions when it does not hold, and also assumes that convergence in probability implies almost sure convergence.

[3] The original presentations of these works contained some errors; Leonenko and Pronzato (2010) provide corrections for some of these theorems.

It is easy to prove that

$$\lim_{\alpha \to 1} R_\alpha(p\|q) = KL(p\|q) \doteq \int p \log \frac{p}{q},$$

where KL stands for the Kullback–Leibler divergence.

**Definition 2.** *Let $p$ and $q$ be densities over $\mathbb{R}^d$. The $L_2$ divergence is defined as*

$$L(p\|q) \doteq \left( \int (p(x) - q(x))^2 \, \mathrm{d}x \right)^{1/2}. \quad (2)$$

In the next section, we will provide consistent estimators for the following quantities:

$$D_\alpha(p\|q) \doteq \int p^\alpha(x) q^{1-\alpha}(x) \, \mathrm{d}x, \quad (3)$$

$$L^2(p\|q) \doteq \int (p(x) - q(x))^2 \, \mathrm{d}x. \quad (4)$$

Plugging these estimates into the appropriate formulas immediately leads to consistent estimators for $R_\alpha(p\|q)$, and $L(p\|q)$.

Now we are ready to formally define the problem. We are given $\{X_{i,1}, \ldots, X_{i,T_i}\}$ i.i.d. samples from $\{f_i\}$ density functions ($i = 1, \ldots, I$, $X_{i,t} \in \mathbb{R}^D$). Later we will also refer to the samples from a single $\{f_i\}$ as a "group". Using these samples, we want to estimate the $L_2$ and Rényi divergences between these $\{f_i\}$ density functions. In the first case $w_{i,j} \doteq \left( \int (f_i(x) - f_j(x))^2 \, \mathrm{d}x \right)^{1/2}$, while in the second case $w_{i,j} \doteq \frac{1}{\alpha-1} \log \int f_i^\alpha(x) f_j^{1-\alpha}(x) \, \mathrm{d}x$. The problem, of course, is that we do not know these $\{f_i\}$ densities, and we want to compute the divergences without estimating them.

Having estimated the $\{w_{i,j}\}$ distances, we can analyze the distributions as if they were points in a finite-dimensional Euclidean space, and $\{w_{i,j}\}$ were the distances between the points. For example, we can cluster the distributions, or embed them into a low-dimensional space while preserving proximity; distributions close to each other should be mapped into points that are also close to each other in the lower dimensional space. This can be done using multidimensional scaling (MDS) (Borg and Groenen, 2005), isomap (Tenenbaum et al., 2000), curvilinear component analysis (Sun et al., 2010) or other methods that require only the pairwise distances. This embedding provides a useful tool for visualization and unsupervised exploration of the data set.

Another interesting application of the proposed divergence estimator is the group anomaly detection problem. We model each group as a bag of features, and assume that the $i$th group has a feature distribution $f_i$. Our goal is to select those groups whose feature distributions are significantly different from the distributions of the other groups. This problem can be addressed using the proposed divergence estimator by first estimating the distances between the groups' feature distributions and then finding those groups that are far away from their neighbors (i.e. groups from low density regions of the space of distributions).

The above mentioned problems are unsupervised in nature. Nonetheless, using the proposed divergence estimators we can also solve the following supervised classification tasks. Let $\{X_{i,1}, \ldots, X_{i,T_i}\}$ be i.i.d. samples from $\{f_i\}$ density functions ($i = 1, \ldots, I$, $X_{i,t} \in \mathbb{R}^D$), and for each $f_i$ distribution there is given a $Y_i$ class label. Our goal is to learn a classifier of the distributions using the $\{(X_{i,1}, \ldots, X_{i,T_i}); Y_i\}$ training data and without estimating the $f_i$ densities.

## 3 The Divergence Estimators

In this section we introduce our estimators for $D_\alpha(p\|q)$, and $L^2(p\|q)$. From now on we will assume that (3) and (4) can be rewritten as

$$D_\alpha(p\|q) = \int_\mathcal{M} \left( \frac{q(x)}{p(x)} \right)^{1-\alpha} p(x) \, \mathrm{d}x, \quad (5)$$

$$L^2(p\|q) = \int_\mathcal{M} (p(x) - 2q(x) + q^2(x)/p(x)) p(x) \, \mathrm{d}x. \quad (6)$$

where $\mathcal{M} = \mathrm{supp}(p)$. Let $X_{1:N} \doteq (X_1, \ldots, X_N)$ be an i.i.d. sample from a distribution with density $p$, and similarly let $Y_{1:M} \doteq (Y_1, \ldots, Y_M)$ be an i.i.d. sample from a distribution having density $q$. Let $\rho_k(x)$ denote the Euclidean distance of the $k$th nearest neighbor of $x$ in the sample $X_{1:N} \setminus x$, and similarly let $\nu_k(x)$ denote the distance of the $k$th nearest neighbor of $x$ in the sample $Y_{1:M} \setminus x$. We will claim that the following estimators are consistent under certain conditions:

$$\widehat{D}_\alpha(X_{1:N}\|Y_{1:M}) \doteq \frac{1}{N} \sum_{n=1}^N \left( \frac{(N-1)\rho_k^d(X_n)}{M\nu_k^d(X_n)} \right)^{1-\alpha} B_{k,\alpha}, \quad (7)$$

where $B_{k,\alpha} \doteq \frac{\Gamma(k)^2}{\Gamma(k-\alpha+1)\Gamma(k+\alpha-1)}$, $k > |\alpha - 1|$.

$$\widehat{L}^2(X_{1:N}\|Y_{1:M}) \doteq$$
$$\frac{1}{N} \sum_{n=1}^N \left[ \frac{k-1}{(N-1)c\rho_k^d(X_n)} - \frac{2(k-1)}{Mc\nu_k^d(X_n)} \right.$$
$$\left. + \frac{(N-1)c\rho_k^d(X_n)}{(Mc\nu_k^d(X_n))^2} \frac{(k-2)(k-1)}{k} \right], \quad (8)$$

where $k - 2 > 0$. The estimators are simple, can avoid the need for density estimation, and use only certain $k$ nearest neighbor statistics with a fixed $k$. In Section 4.2 we will provide an intuitive explanation of the forms of these estimators.

Let $p$, $q$ be bounded away from zero, bounded from above, and uniformly continuous density functions. Let $\mathcal{M} = \text{supp}(p)$ be a finite union of bounded convex sets. We have the following main theorems.

**Theorem 3** ($L_2$ divergence estimator). *Under the conditions listed above, $\lim_{N,M\to\infty} \mathbb{E}[\widehat{L}^2] = L^2$, and $\lim_{N,M\to\infty} \mathbb{E}\left[(\widehat{L}^2 - L^2)^2\right] = 0$, i.e., the estimator is asymptotically unbiased and $L_2$ consistent.*

**Theorem 4** (Rényi divergence estimator). *Let $k > 2|\alpha - 1|$. Under the conditions listed above we have that $\lim_{N,M\to\infty} \mathbb{E}[\widehat{D}_\alpha] = D_\alpha$ and $\lim_{N,M\to\infty} \mathbb{E}\left[(\widehat{D}_\alpha - D_\alpha)^2\right] = 0$, i.e., the estimator is asymptotically unbiased and $L_2$ consistent.*

## 4 Consistency Proofs

Due to the lack of space, we provide only a brief sketch of the proofs. The detailed proofs will be published elsewhere. In the Supplementary material (Póczos et al., 2011), we demonstrate the consistency of the estimators by some numerical experiments as well.

### 4.1 $k$-NN Based Density Estimators

We will exploit some properties of $k$-NN based density estimators. In this section we define these estimators and briefly summarize their most important properties.

$k$-NN density estimators operate using only distances between the observations in a given sample ($X_{1:N}$, or $Y_{1:M}$) and their $k$th nearest neighbors. Loftsgaarden and Quesenberry (1965) define the $k$-NN based density estimators of $p$ and $q$ at $x$ as follows.

**Definition 5** ($k$-NN based density estimators).

$$\hat{p}_k(x) = \frac{k/N}{\mathcal{V}(\mathcal{B}(x, \rho_k(x)))} = \frac{k}{Nc\rho_k^d(x)}, \quad (9)$$

$$\hat{q}_k(x) = \frac{k/M}{\mathcal{V}(\mathcal{B}(x, \nu_k(x)))} = \frac{k}{Mc\nu_k^d(x)}. \quad (10)$$

*Here $\mathcal{B}(x, R)$ denotes a closed ball around $x \in \mathbb{R}^d$ with radius $R$, and $\mathcal{V}(\mathcal{B}(x, \nu_k(x))) = cR^d$ is its volume where $c$ stands for the volume of a $d$-dimensional unit ball.*

The following theorems show the consistency of these density estimators.

**Theorem 6** ($k$-NN density estimators, convergence in probability). *If $k(N)$ denotes the number of neighbors applied at sample size $N$, $\lim_{N\to\infty} k(N) = \infty$, and $\lim_{N\to\infty} N/k(N) = \infty$, then $\hat{p}_{k(N)}(x) \to_p p(x)$ for almost all $x$.*

**Theorem 7** ($k$-NN density estimators, almost sure convergence in sup norm). *If $\lim_{N\to\infty} k(N)/\log(N) = \infty$ and $\lim_{N\to\infty} N/k(N) = \infty$, then $\lim_{N\to\infty} \sup_x |\hat{p}_{k(N)}(x) - p(x)| = 0$ almost surely.*

Note that these estimators are consistent only when $k(N) \to \infty$. We will use these density estimators in our proposed divergence estimators; however, we will keep $k$ fixed and will still be able to prove their consistency.

### 4.2 Proof Outline for Theorems 3-4

**Lemma 8** (Lebesgue (1910)). *If $\mathbb{R}^d \supset \mathcal{M}$ is a Lebesgue measurable set, and $g \in L_1(\mathcal{M})$, then for any sequence of $R_n \to 0$, $\delta > 0$, and for almost all $x \in \mathcal{M}$, there exists an $n_0(x,\delta) \in \mathbb{Z}^+$ such that if $n > n_0(x,\delta)$, then*

$$g(x) - \delta < \frac{\int_{\mathcal{B}(x,R_n)} g(t)\, \mathrm{d}t}{\mathcal{V}(\mathcal{B}(x,R_n))} < g(x) + \delta. \quad (11)$$

We can see from (9) that the $k$-NN estimation of $1/p(x)$ is simply $Nc\rho_k^d(x)/k$. Using the Lebesgue lemma (Lemma 8), one can prove that the distribution of $Nc\rho_k^d(x)$ converges weakly to an Erlang distribution with mean $k/p(x)$, and variance $k/p^2(x)$ (see Leonenko et al. (2008) for the details). Therefore, if we divide $Nc\rho_k^d(x)$ by $k$, then asymptotically it has mean $1/p(x)$ and variance $1/(kp^2(x))$. It implies that indeed (in accordance with Theorems 6–7) $k$ should converge to infinity in order to get a consistent density estimator, otherwise the variance will not disappear. On the other hand, $k$ cannot grow too fast: if say $k = N$, then the estimator would be simply $c\rho_k^d(x)$, which is a useless estimator since it is asymptotically zero whenever $x \in \text{supp}(p)$.

Luckily, in our case we do not need to apply consistent density estimators. The trick is that (5)–(6) have special forms; each term inside these equations has $\int p(x)p^\gamma(x)q^\beta(x)\,\mathrm{d}x$ form. In (7)–(8), each of these terms is estimated by

$$\frac{1}{N}\sum_{i=1}^{N}\left(\hat{p}_k(X_i)\right)^\gamma \left(\hat{q}_k(X_i)\right)^\beta B_{k,\gamma,\beta}, \quad (12)$$

where $B_{k,\gamma,\beta}$ is a correction factor that ensures asymptotic unbiasedness. The value of $B_{k,\gamma,\beta}$ correction terms can be determined using the following argument. By Lemma 8, we can prove that the distributions of $\hat{p}_k(X_i)$ and $\hat{q}_k(X_i)$ converge weakly to the Erlang distribution with means $k/p(X_i)$, $k/q(X_i)$ and variances $k/p^2(X_i)$, $k/q^2(X_i)$, respectively (Leonenko et al., 2008). Furthermore, they are conditionally independent for a given $X_i$. Therefore, "in the limit" (12) is simply the empirical average of the products of the $\gamma$th (and $\beta$th) powers of independent Erlang distributed variables. These moments can be calculated using Lemma 9 below:

**Lemma 9** (Moments of the Erlang distribution). *Let $f_{x,k}(u) \doteq \frac{1}{\Gamma(k)} \lambda^k(x) u^{k-1} \exp(-\lambda(x) u)$ be the density of the Erlang distribution with parameters $\lambda(x) > 0$ and $k \in \mathbb{Z}^+$. Let $\gamma \in \mathbb{R}$ such that $\gamma + k > 0$. Then the $\gamma$th moments of this Erlang distribution can be calculated as $\int_0^\infty u^\gamma f_{x,k}(u) \, du = \lambda(x)^{-\gamma} \frac{\Gamma(k+\gamma)}{\Gamma(k)}$.*

With the help of this lemma we can set $B_{k,\gamma,\beta}$ to a value with which $(\hat{p}_k(X_i))^\gamma (\hat{q}_k(X_i))^\beta B_{k,\gamma,\beta}$ is an asymptotically unbiased estimator for $\int p(x) p^\gamma(x) q^\beta(x) \, dx$ and has bounded variance. Note that for a fixed $k$, the $k$-NN density estimator is not consistent since its variance does not vanish. In our case, however, this variance will disappear thanks to the empirical average in (12) and the law of large numbers.

While the underlying ideas of this proof are remarkably simple, there are a couple of serious gaps in it. Most importantly, from the Lebesgue lemma (Lemma 8) we can guarantee only the weak convergence of $\hat{p}_k(X_i)$, $\hat{q}_k(X_i)$ to the Erlang distribution. From this weak convergence we cannot imply that the moments of the random variables converge too. To handle this issue, we will need stronger tools such as the concept of asymptotically uniformly integrable random variables (van der Wart, 2007), and we also need the uniform generalization of the Lebesgue lemma. As a result, in Theorems 3–4 we need to put some extra conditions on the densities $p$ and $q$ (such as they should be bounded away from zero, bounded above, uniformly continuous, and $\mathcal{M}$ should be a finite union of bounded convex sets).

## 5 Experiments

In this section, we test the performance of the proposed divergence estimators. In these experiments, we first estimate the divergences and then use them for tasks including embedding, clustering, classification, and group anomaly detection.

We mainly compare our nonparametric (*NP*) methods to the alternative in which we first fit a parametric Gaussian mixture model (GMM) to each group, and then calculate the divergences between these GMMs. This alternative method is called *GMM estimation*. An even simpler method is to fit a single Gaussian to each group and then calculate the divergences between these Gaussians. We call this method *Gaussian estimation*. For GMMs, we have the analytical result for their $L_2$ divergences (Jian and Vemuri, 2005). However, there is no closed formula for the Rényi divergence between two GMMs, hence we resort to MCMC methods to approximate this quantity. In all of our experiments we use $\alpha = 0.5$ for the Rényi divergence. We also symmetrize both the Rényi and $L_2$ divergences by taking the average of the two way estimations: $D_{sym}(p|q) = (D(p\|q) + D(q\|p))/2$.

### 5.1 Embedding of Distributions

In this experiment we use synthetic data to demonstrate how the proposed estimators can be used to embed simple distributions including uniform, beta, and 1-dimensional Gaussian. For each type of distribution, we realize many distributions using different parameters, and each realization generates a group of data. Then we estimate divergences between these groups and embed them into a low-dimensional space using multidimensional scaling (Borg and Groenen, 2005). Finally, we visualize the embedded groups and see if the underlying structure of parameters are captured by the embedding.

For uniform and Gaussian distributions, the parameters are the mean and standard deviation. We selected the parameters from a uniform $10 \times 10$ grid, where the mean and standard deviation vary within $[0, 1]$ and $[0.3, 0.7]$ respectively. For beta distributions, we use the canonical parametrization with parameters $\alpha, \beta$, and select their values from a uniform $10 \times 10$ grid on $[0.7, 3] \times [0.7, 3]$. To visualize the results, we color the embedded groups according to the above parameters. For each group we generate $2\,000$ samples. For the nonparametric estimators we use $k = 20$ nearest neighbors.

We compare our NP estimators to the Gaussian estimations. As the ground truth, we also calculate the embedding using the true divergences between the underlying distributions. Results using both the $L_2$ divergence (Figure 1 (a)-(c)) and Rényi divergence (Figure 1 (d)-(f)) are shown. We can see that the NP estimator can reveal the structure of the underlying parameters, and always produces embeddings that are similar to the ground truth. On the other hand, the embeddings by simple Gaussian estimation can be quite poor when the distribution is very different from Gaussian.

Next we show how embedding can reveal the structure of more complex distributions. To generate the data for groups, we first uniformly sample $3\,000$ points from sine curves $y = sin(\theta x)$, where $x \in [0, 2\pi]$, and $\theta$ is selected uniformly over $[2, 4]$. Then we added Gaussian noise from $\mathcal{N}(0, 0.3^2)$ to each $(x, y)$ pair we sampled. Two groups of data are shown in Figure 2.

We embedded the groups into a 2D space using the proposed estimators with $k = 20$ nearest neighbors. We also performed the embedding using GMM estimations. Results using both methods are shown in Figure 3. The nonparametric approach correctly reveals the 1-

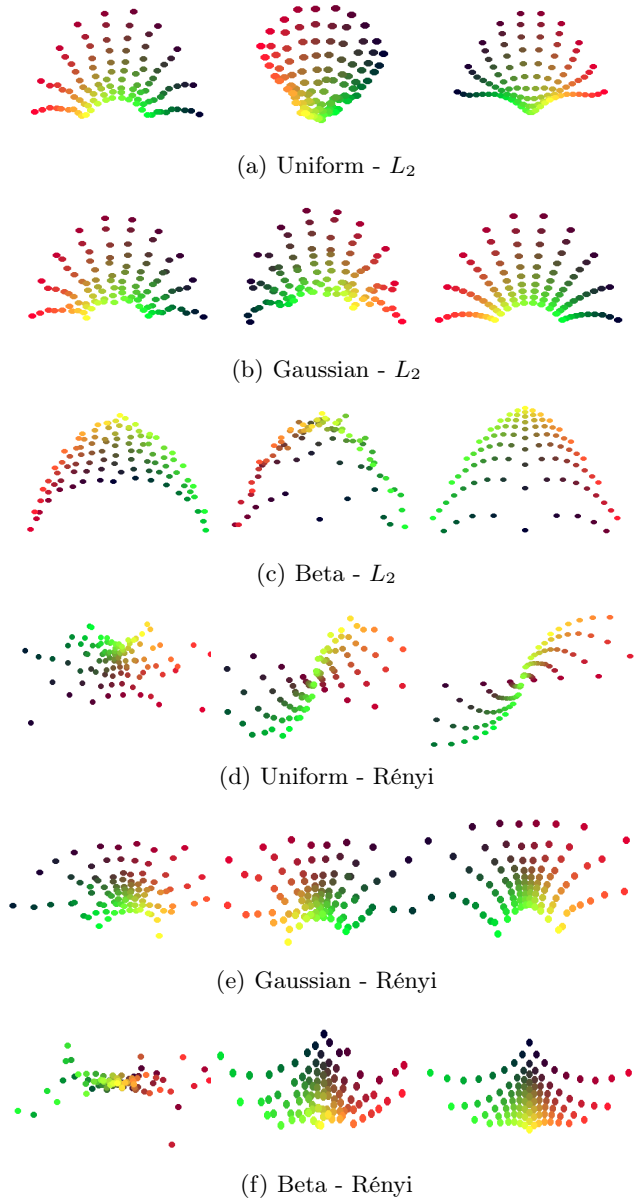

Figure 1: (a)-(c): Embeddings of Uniform, Gaussian and Beta distributions using $L_2$ divergence. (d)-(f): Results using Rényi divergence. Gaussian and Uniform distributions are colored by their means and variances (the red and green color component respectively). Beta distributions are colored by the two parameters. From left to right, the embeddings are produced by the Gaussian estimation, NP estimation, and the true divergence.

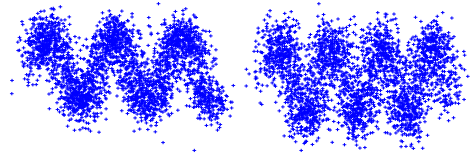

Figure 2: Two groups of the simulated noisy sine data.

dimensional nature of these distributions and orders the groups by their frequency. On the other hand, GMM estimation fails when the number of components is small. Although with enough components GMM can eventually work, it involves excessive computation and parameter tuning that are not needed in NP estimators.

### 5.2 Image Clustering and Classification

We can also use the proposed estimators to facilitate clustering and classification tasks by feeding the estimated divergences to algorithms that only need the dissimilarities between instances. In our experiments we use $k$-nearest-neighbors ($k$-NN) based clustering algorithms and classifiers.

We test the performance on the image data from Fei-Fei and Perona (2005). We adopt the "bag-of-words" representation for the images. Each image is a group of local patches, and each patch has a feature vector. We assume that each image has an inherent distribution to generate its patches, and these patches are i.i.d. In other words, each image is a distribution, and its patches are samples from this distribution. Then, we can measure the dissimilarity between images by estimating the divergences between the corresponding distributions. Note that we do *not* quantize the patches as in Fei-Fei and Perona (2005), but rather use the original real-valued features and deal with their distributions directly.

Specifically, we use the categories "MITmountain", "MITcoast", "MIThighway", and "MITinsidecity" from the data set. From each category we randomly select 50 images. Features are extracted as in Fei-Fei and Perona (2005). Points are firstly sampled on a uniform grid with interval 5. At each point, we extract the 128-dimensional SIFT features (Lowe, 2004) and then reduce its dimension to 2 using PCA. In the end, we have 200 groups (images), each of which contains about 1 600 2-dimensional points (patches).

We compare the NP estimators to the Gaussian estimator and the 5-component-GMM estimator. We also compare them to the algorithm described by Bosch et al. (2006), which reflects the performance of a conventional "bag-of-word" (BoW) approach. In this BoW method, patches are quantized to 100 "visual words", and each image is represented as a 100-dimensional histogram of these words. Then *probabilistic latent semantic analysis* (pLSA) by Hofmann (1999) is applied to embed the images into a latent semantic space to get low-dimensional representations called topic distributions (here 20 topics are used, and thus each image is converted to a 20-dimensional probability vector). Finally, Euclidean distances between these topic distributions are used to measure the dissimilarities between images.

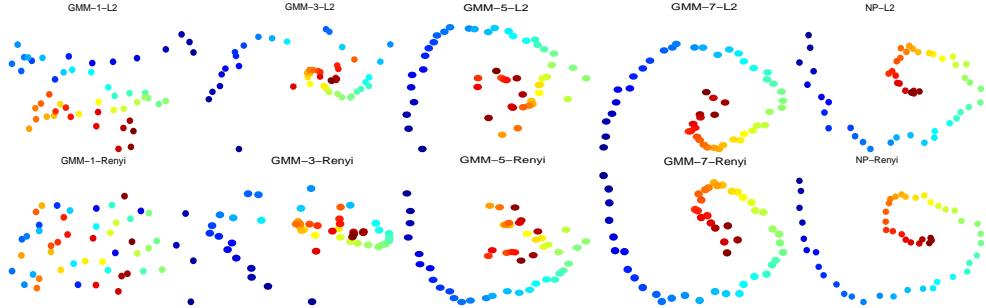

Figure 3: Embeddings of the 2-dimensional noisy sine data using GMMs and our nonparametric estimators. Points are colored by their underlying frequency. Column 1 to 4 show the embeddings by GMMs with increasing number of Gaussian components. The last column shows the embedding by our nonparametric estimators. The first row uses the $L_2$ divergence and the second row uses the Rényi divergence.

To cluster these images, we feed the divergences to the *spectral clustering* algorithm by Zelnik-Manor and Perona (2004). To evaluate the clustering results, we first form a confusion matrix from the category labels and the cluster labels, then permute the columns to maximize the trace of this matrix, which is equal to the number of correctly identified groups. We repeat 20 random runs and report the results in Figure 4(a).

The first thing we can observe is that the Rényi divergence performs better than the $L_2$ divergence for this data set. We can also see that the Gaussian estimator is clearly inadequate. The GMM estimator improves over the single Gaussian one but is still slightly worse than the NP. The standard BoW approach also produces a slightly worse results than NP. Paired t-tests show that the difference between GMM and NP is significant, but the difference between NP and BoW is not significant (p-value is $6 \times 10^{-3}$ for GMM-Rényi vs. NP-Rényi, and 0.94 for NP-Rényi vs. BoW).

We can also use the divergences for classification of distributions. Here we adopt a simple $k$-NN strategy: a group's label is predicted based on votes from the labeled groups that are closest, i.e., have the smallest divergence. We use $k = 11$ nearest neighbors for this classifier. In each run, we conduct 10-fold cross-validation on the randomly selected images and report the classification accuracy. The results from 20 random runs are reported in Figure 4(b). Similar results can be observed as in the clustering task; the nonparametric Rényi divergence estimator achieves the best performance among the competitors. Paired t-test gives p-value $5.33 \times 10^{-4}$ for the difference between GMM-Rényi and NP-Rényi, and 0.15 for NP-Rényi vs. BoW. We also note that the nonparametric $L_2$ estimators produced poor results in this experiment. In the Supplementary material (Póczos et al., 2011) we show that for small sample size the $L_2$ estimator might have larger bias and variance than the that of the Rényi divergence estimator; this can result in poor performance.

### 5.3 Group Anomaly Detection

One novel application of our divergence estimators is the detection of anomalous groups of data points. Note that unlike traditional anomaly detection methods that focus on unusual points, a group may have an anomalous distribution of points even if none of the individual points are unusual.

We use a simple detection algorithm based on nearest neighbors (Zhao and Saligrama, 2009). In this case the anomaly score of a group (distribution) is just the divergence between this group and its $k$th nearest neighbors. We apply this detector and compare the performance of different divergence estimators. The performance is measured by the *area under the ROC curve* (AUC).

We experimented with the different divergence estimators on the images we used before. The normal data was defined to be images from the categories in the previous experiment. In addition, we used images from two other categories "MITforest" and "livingroom" as anomalies. We used a random 75% of the normal images as training data and the rest 25% for testing. We also add some anomalous images to the test set to make it half normal and half abnormal. Then we asked the anomaly detector to find the anomalies, i.e., "livingroom" and "forest" images from this mixture. Those test groups that were the furthest away from their nearest neighbors in the training set were selected as anomalous groups.

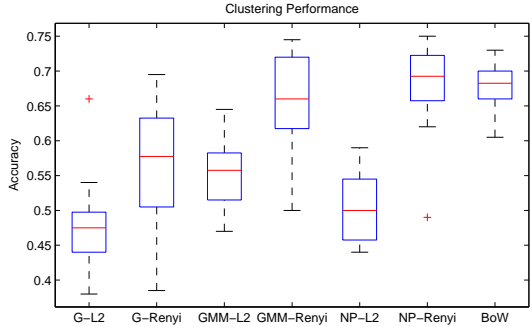

(a) Spectral clustering performance.

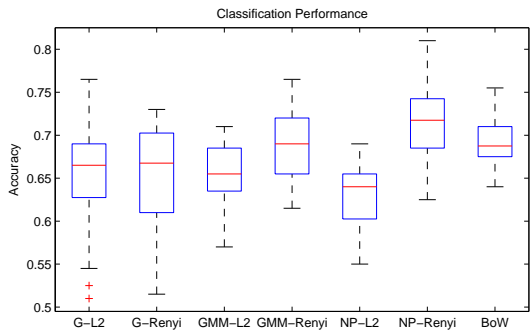

(b) $k$-NN classification performance.

Figure 4: Clustering and classification performance using different divergence estimators. The columns correspond to the Gaussian estimation (G), GMM estimation, and nonparametric divergence estimation (NP) using Rényi and $L_2$ divergences, respectively. Finally, the column on the right displays the performance of the BoW method.

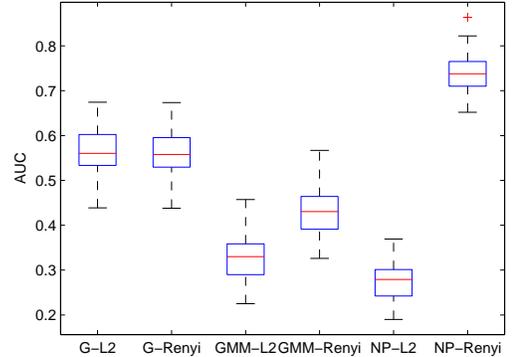

Figure 5: Anomalous image detection performance.

We again compare the NP, Gaussian, and GMM estimators in this task. We use 5 Gaussian components in the GMM. The anomaly score of a test group is the divergence between the group and its 5th nearest neighbor in the training set. The results from 100 random runs are shown in Figure 5. Our NP estimator for Rényi divergence produces the best results, and the $L_2$ divergence again performs poorly. It is also interesting to see that the GMM estimator failed as well. Various reasons can cause this result, e.g., the inherent difference between normal and abnormal images can influence the estimators, and the GMM may overfit the data.

In the next experiment, we detect anomalous galaxy clusters in the astronomical data set from Sloan Digital Sky Survey[4] (SDSS). SDSS contains about $7 \times 10^5$ galaxies, each of which has a 4 000-dimensional continuum of the spectrum. We downsampled the continuum to get a 500-dimensional feature vector for each galaxy.

The "friends-of-friends" method (Garcia, 1993) was used to find spatial clusters (groups of nearby galaxies). 505 groups (7 530 galaxies) were found, each of which contains about 10–50 galaxies. In each group we used PCA to reduce the 500-dimensional continuum to 2-dimensional features preserving 95% of the variance. Note that this data set could be difficult for the NP estimators since the group sizes are small.

Due to the lack of labels, we use artificially injected anomalies to get statistically meaningful results. These injected groups are synthesized in the way such that each group consists of normal galaxies, but the distribution of the galaxies' features are rare in real galaxy clusters. In each run we injected 10 such random anomalies, and the whole data set contained 515 groups.

The AUC results from 20 random runs are shown in Figure 6. In this problem, the NP $L_2$ estimator achieves the best performance, and the NP estimators clearly outperform the parametric alternatives.

## 6 Discussion and Conclusion

We developed a new framework and proposed algorithms for several machine learning problems performed on the space of distributions. These problems include low dimensional embedding, clustering, classification and outlier/anomaly detection. Most of the machine learning algorithms operate on fixed finite dimensional feature representation. Kernel methods might transform the instances temporarily to an infinite dimensional space, but the ultimate goal is still the same: to solve the classification, clustering, outlier detection, low dimensional embedding problems in the original finite dimensional feature space. In our setting, the space of our features (continuous distributions) is infinite dimensional. Furthermore, in contrast to standard machine learning problems we cannot observe them directly, only a few i.i.d. samples are available for us to represent these distributions.

---

[4] http://www.sdss.org

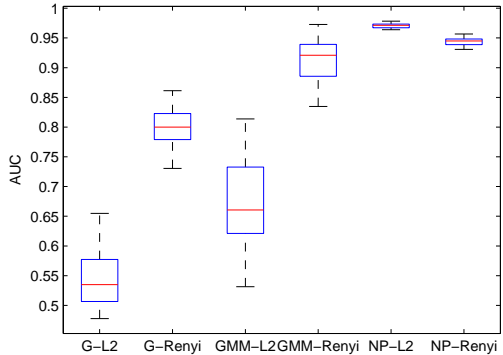

Figure 6: Anomalous galaxy cluster detection performance.

This new framework has many potential applications from bioinformatics to astronomy. It is useful anywhere where we take measurements of objects and our goal is to differentiate the divergences between the distributions of these measurements. We demonstrated the applicability of our framework both on synthetic toy problems and on real world problems including computer vision and anomaly detection in astronomical data.

In this paper we used nonparametric Rényi and $L_2$ divergence estimators to estimate the deviation between distributions. We provided a brief sketch for the proof of their consistency. The technical details will be published elsewhere. We also compared our nonparametric estimators with a few competitors including a parametric estimator that assumes the distributions to be Gaussians, and a more complex estimator that first fits a mixture of Gaussian to the data and then estimates the divergences between these mixtures.

We found that our nonparametric estimators outperform the competitors under various conditions. If the data does not match the parametric assumptions, then parametric approaches can lead to poor divergence estimators. Even though many distributions can be well-approximated by mixture of Gaussians, this approach might be too slow and sensitive to the number of Gaussian components in the model. The $L_2$ divergence can be easily calculated between two mixtures of Gaussians; however, it is challenging to calculate the Rényi divergence between them, and we might need Monte Carlo methods to approximate this divergence.

Empirically we observed that the $L_2$ and Rényi estimators exhibit different behaviors and their performances depend on the actual distributions. We also found that the Rényi divergence is usually easier to estimate well than the $L_2$ divergence, which seems to be more sensitive to outliers.

There are several open questions left waiting for answers. Currently, the convergence rates of our divergence estimators are unknown. It would also be desirable to derive theoretical bounds on the sample complexity for many of these machine learning tasks defined on distributions.